\documentclass[runningheads]{llncs}
\usepackage[T1]{fontenc}
\usepackage{lmodern}
\usepackage{amsmath,amssymb,booktabs,graphicx,xcolor}
\usepackage{url}
\usepackage{tikz}
\usetikzlibrary{arrows.meta}
\newcommand{\Deltaauth}{\Delta_{\mathrm{authority}}}
\newcommand{\asel}{a_{\mathrm{sel}}}
\newcommand{\sys}{AuthEval}
\begin{document}
\title{Authority-Preserving Evaluation of \\ Medical Vision-Language Assistants}
% uncomment below for including orcid
% \author{Flint Xiaofeng Fan\inst{1,2}\orcidID{0000-0003-1658-4699} \and
% Cheston Tan\inst{2}\orcidID{0000-0003-1248-4906} \and
% Yew-Soon Ong\inst{2}\orcidID{0000-0002-4480-169X} \and
% Roger Wattenhofer\inst{1}\orcidID{0000-0002-6339-3134}}

\author{Flint Xiaofeng Fan\inst{1,2} \and
Cheston Tan\inst{2} \and
Yew-Soon Ong\inst{2} \and
Roger Wattenhofer\inst{1}}
% %
\authorrunning{F. X. Fan et al.}
% \institute{
% ETH Zurich, Zurich, Switzerland
% \email{\{xiafan,wattenhofer\}@ethz.ch}
% \and
% Agency for Science, Technology and Research (A*STAR), Singapore, Singapore\\
% \email{\{fanx,cheston-tan,ong_yew_soon\}@a-star.edu.sg}}

\institute{
ETH Zurich, Zurich, Switzerland\\
\email{\{xiafan,wattenhofer\}@ethz.ch}
\and
Agency for Science, Technology and Research (A*STAR), Singapore, Singapore\\
\email{\{fanx,cheston-tan,ong\_yew\_soon\}@a-star.edu.sg}}
% \titlerunning{FedPref for Structured Radiology Extraction}

%
\maketitle              % typeset the header of the contribution
\begin{abstract}
Medical vision-language models can propose how urgently a skin lesion should be reviewed,
but the local service retains authority to accept or replace that proposal under referral
policy, capacity, and locally held patient context. Proposal quality and selected-action
quality are therefore distinct evaluation targets, and benchmark evidence transfers between
them only when local review preserves the expected action score. We introduce \sys{}, a
logging and evaluation framework that records both actions, scores the selected action under
declared local criteria, and, where feasible, scores the declined proposal under the same rule.
It reports the resulting authority gap only when the record supports it. Because the gap is
the product of the proposal-change rate and the mean score change on changed cases, that rate
alone determines neither its magnitude nor its sign. On ISIC 2019, with MedGemma and simulated
local review, two constraint regimes with similar change rates produced an optimistic
image-equal gap under capacity ($+0.744$ simulator units) but no detectable gap under safety.
The declared evaluation unit also mattered: under mixed constraints the gap reversed from
$+0.374$ to $-0.206$ when weighting shifted from image to lesion-aware cluster. \sys{} thus
clarifies whether a study's records support claims about the model, the workflow, or both.
\end{abstract}

\section{Introduction}

A lesion image enters a teledermatology pathway, and a medical vision-language model
(VLM) proposes how urgently it should be reviewed. The local service, however,
retains authority over the routing decision: referral policy, available capacity,
follow-up safeguards, and clinician review may lead it to accept or replace the
proposal~\cite{nice2025skinai,marsden2024teledermatology}. The action retained by the
workflow is therefore the one the service selected, which need not be the one the model
proposed.

Medical VLM benchmarks evaluate model outputs against
references~\cite{royer2024multimedeval,liu2025medvlmbenchmark}, addressing proposal quality.
Workflow-level claims additionally require selected-action quality because local review may
accept or replace the proposal. This changes the evaluation
construct~\cite{raji2021benchmark}: replacement can either improve or worsen the score under
local criteria.

\begin{figure}[t]
\centering
\resizebox{\linewidth}{!}{%
\def\lesioniconB{\tikz[baseline=-0.55ex,scale=.34]{%
  \draw[semithick,blue!55!black] (0,0) circle (.43);
  \fill[brown!55!black]
    plot[smooth cycle,tension=.75] coordinates {(-.25,.06) (-.12,.27) (.13,.24)
      (.29,.04) (.18,-.22) (-.08,-.27) (-.28,-.10)};}}
\def\lockiconB{\tikz[baseline=-0.55ex,scale=.34]{%
  \draw[semithick,rounded corners=.5pt] (-.42,-.34) rectangle (.42,.18);
  \draw[semithick] (-.27,.18) arc[start angle=180,end angle=0,radius=.27];
  \fill[black!60] (0,-.05) circle (.055);}}
\def\gateiconB{\tikz[baseline=-0.55ex,scale=.35]{%
  \draw[semithick,green!40!black] (-.52,.34)--(.52,.34)--(.18,-.05)--(.18,-.35)--(-.18,-.35)--(-.18,-.05)--cycle;}}
\newcommand{\icoB}[1]{\makebox[14pt][c]{\raisebox{0pt}[6.8pt][1.8pt]{#1}}}
\begin{tikzpicture}[
  >={Latex[length=1.9mm]}, font=\small,
  boxB/.style={rounded corners=2pt, semithick, align=center, inner sep=2.2pt,
              text width=18mm, minimum height=11mm},
  moduleB/.style={boxB, draw=blue!55!black, fill=blue!8},
  gateB/.style={boxB, draw=green!45!black, fill=green!12},
  actionB/.style={boxB, draw=green!45!black, fill=green!18},
  yB/.style={rounded corners=2pt, semithick, align=center, inner sep=2.2pt,
              text width=20mm, minimum height=7mm, draw=black!55, dashed, fill=black!3},
  percaseTop/.style={boxB, text width=20mm, draw=orange!95!black, dashed, fill=orange!10},
  percaseBot/.style={boxB, text width=20mm, draw=green!45!black, fill=green!13},
  aggTop/.style={boxB, text width=16mm, draw=orange!95!black, fill=orange!15},
  aggBot/.style={boxB, text width=16mm, draw=green!45!black, fill=green!18},
  gapB/.style={boxB, text width=26mm, semithick, draw=black!70, fill=black!4},
  flowB/.style={->, semithick},
  ctxB/.style={->, densely dotted, semithick, draw=black!65},
  subB/.style={font=\footnotesize},
]
\def\xA{0} \def\xB{2.05} \def\xC{4.1} \def\xD{6.15}
\def\xE{8.9} \def\xF{11.75}
\def\yA{0} \def\yB{-1.85} \def\yC{-3.55}

% Panel headers and divider.
\node[font=\bfseries,text=blue!45!black] at (\xB+0.55,2.05) {Decision time};
\draw[black!60] (\xA-0.85,1.7) -- (7.4,1.7);
\node[font=\bfseries,text=black!55] at (\xE+0.55,2.05) {Post-hoc evaluation};
\draw[black!60] (7.75,1.7) -- (\xF+1.4,1.7);
\draw[black!35] (7.58,1.85) -- (7.58,\yB-0.95);

% Decision-time pipeline (all blue: proposal generation and its output).
\node[moduleB] (xB) at (\xA,\yA) {\icoB{\lesioniconB}\quad $x$\\[-1pt]{\footnotesize case input}};
\node[moduleB] (modelB) at (\xB,\yA) {VLM + action readout};
\node[moduleB] (slateB) at (\xC,\yA) {$S$\\[-1pt]{\footnotesize ranked slate}};
\node[moduleB] (skipB) at (\xD,\yA) {$\hat a$\\[-1pt]{\footnotesize proposed action}};
\draw[flowB, blue!55!black] (xB)--(modelB);
\draw[flowB, blue!55!black] (modelB)--(slateB);
\draw[flowB, blue!55!black] (slateB)--(skipB);

% Local review path.
\node[boxB, draw=black!55, dashed, fill=black!3] (zB) at (\xA,\yB) {\icoB{\lockiconB}\quad $z$\\[-1pt]{\footnotesize local context}};
\node[gateB] (gB) at (\xC,\yB) {\icoB{\gateiconB}\quad $G(x,z,S)$\\[-1pt]{\footnotesize local review}};
\node[actionB] (selB) at (\xD,\yB) {$\asel$\\[-1pt]{\footnotesize selected action}};
\draw[flowB, blue!55!black] (xB.south) -- ++(0,-.25) -| ([xshift=-8mm]gB.north);
\draw[ctxB] (zB)--(gB);
\draw[flowB, green!45!black] (slateB)--(gB);
\draw[flowB, green!45!black] (gB)--(selB);

% Shared post-hoc scoring rule: a compact y arrow, plus a frame receiving z too.
\node[font=\small, text=black!60] (yBn) at (\xE,1.35) {reference $y$};
\draw[draw=black!45, dashed, rounded corners=3pt, fill=black!3]
  (\xE-1.2,0.78) rectangle (\xE+1.2,\yB-0.7);
\draw[flowB, black!55] (yBn.south)--(\xE,0.78);
\node[subB, text=black!75] at (\xE,0.5) {$U_z(x,y,a)$};
\node[percaseTop] (upropB) at (\xE,\yA-0.55) {$U_z(x,y,\hat a)$\\[-1pt]{\footnotesize extra field when $\hat a\neq\asel$}};
\node[percaseBot] (uselB) at (\xE,\yB) {$U_z(x,y,\asel)$\\[-1pt]{\footnotesize per case}};
\draw[flowB, orange!95!black] (skipB.east)--(upropB.west);
\draw[flowB, green!45!black] (selB.east)--(uselB.west);
\draw[ctxB] (zB.south) -- ++(0,-0.65) -| (\xE-1.2,\yB-0.35);

% Aggregation and population scores.
\node[aggTop] (jauthB) at (\xF,\yA) {$J_{\rm auth}$\\[-1pt]{\footnotesize proposal score}};
\node[aggBot] (jB) at (\xF,\yB) {$J$\\[-1pt]{\footnotesize selected-action score}};
\draw[flowB, orange!95!black] (upropB.east)--(jauthB.west);
\draw[flowB, green!45!black] (uselB.east)--(jB.west);

% Authority gap: centered beneath both population scores, two short symmetric drops.
\node[gapB] (gapB2) at (\xF+2.5,{(\yA+\yB)/2}) {\textbf{authority gap}\\[-1pt]$\Deltaauth=J_{\rm auth}-J$};
\draw[flowB, orange!95!black] (jauthB.east) -| (gapB2.north);
\draw[flowB, green!45!black] (jB.east) -| (gapB2.south);
\end{tikzpicture}%
}
\caption{Why proposal and selected-action scores can differ, organized by \emph{when}
information is available. At decision time the proposer sees only $x$, while the label-blind
gate $G$ also sees local context $z$ and slate $S$, but never the reference $y$. Post-hoc
scoring applies one locally parameterized rule to both actions; the declined proposal's score
is the one extra field needed when review changes the action.}
\label{fig:overview}
\end{figure}
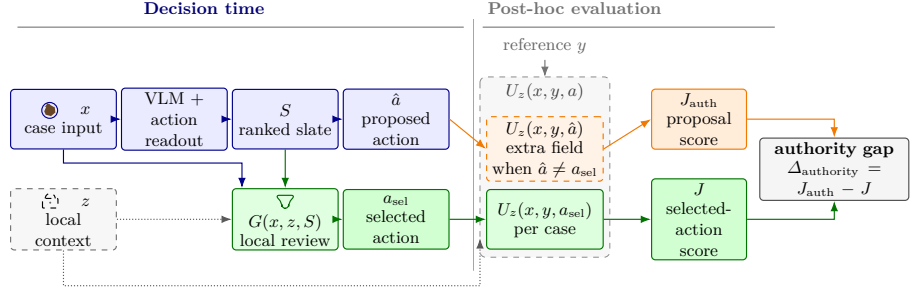

A role-separated log with both actions and the selected-action score reveals
how often local review changes the proposal and how well the selected actions score.
It is still insufficient to measure the authority gap: on changed
cases, the score that the declined proposal would have received under the same local
criteria remains unknown. Because this score may depend on locally held context, it
must be computed where that context is available. Figure~\ref{fig:overview}
summarizes the two evaluation paths and this boundary.

This study asks which recorded fields suffice to recover the authority gap, whether the
frequency of local change determines it, and how it varies with the review process, the
proposal mechanism, and the declared evaluation unit. We answer these with \sys{}, a
logging protocol and evaluator that records the proposed and selected actions, scores both
under the same local criteria when feasible, and otherwise marks the gap as not recoverable
rather than treating proposal quality as workflow quality. Factorizing the gap into the
proposal-change rate and the mean score change on changed cases shows why change frequency
alone cannot recover it, and whether proposal scoring is optimistic or pessimistic relative
to selection. In a controlled ISIC 2019 study the gap varies across all three dimensions and
reverses sign between image- and cluster-equal weighting.

\section{Related Work}
\label{sec:related}

Medical VLM benchmarks evaluate model outputs across a range of tasks
~\cite{royer2024multimedeval,liu2025medvlmbenchmark}, but a different question arises when a
study seeks evidence about the combined workflow, in which a local service may accept
or replace the model's proposal. \sys{} complements proposal evaluation by
asking whether proposal-level evidence remains representative after local
selection.

Clinical alert studies distinguish appropriateness from provider
response~\cite{mccoy2012alerts}. Human--AI research separates advice, privately held
information, and final decisions~\cite{bordt2022private} and distinguishes reliance frequency
from its payoff~\cite{guo2024reliance}, while learning-to-defer methods optimize which actor
decides~\cite{mozannar2020defer}. \sys{} addresses the complementary measurement problem of
determining which recorded fields suffice to recover the signed proposal-selection score
difference under the same local criteria.

Federated learning similarly exploits decentralized evidence while keeping raw observations local, including in sequential decision-making settings such as federated Bayesian Optimization and Reinforcement Learning~\cite{dai2024federated}. FedPref learns from local rankings of model-proposed radiology extractions~\cite{fan2026fedpref}, while FedRLHF uses client-held feedback for collaborative policy learning without pooling raw feedback~\cite{fan2025fedrlhf}.
AuthEval complements these learning settings by measuring how local evidence and constraints alter the selected action and its score.

\section{Authority-Preserving Evaluation (\sys{})}
\label{sec:framework}

We call an evaluation \emph{authority-preserving} when its record keeps the model's proposal
and the accountable local selection as distinct objects scored under the same declared rule. The term describes
the evaluation record, not the clinic: \sys{} neither allocates clinical authority nor
certifies that a local selection was appropriate.

A case provides proposer-visible information $x$ and locally held clinical or operational context
$z$, together with action set $\mathcal A$. A medical VLM and declared action readout form a
policy $\pi$ that produces a ranked list of options, or slate, $S_\pi(x)$. Its first option is the
proposal $\hat a$. A label-blind local review rule, gate $G$, selects
$\asel=G(x,z,S_\pi(x))$ and may accept the proposal or substitute another listed action. Post-hoc
evaluation additionally uses a fixed reference state $y$, a diagnosis-derived risk target used
only for scoring. The gate does not observe $y$, so its selection cannot use the
evaluation target.

Let $U_z(x,y,a)$ be a declared post-hoc action score, with larger values preferred. It scores
either action using the same local criteria and fixed reference $y$, and all expectations use
the declared weighting rule. Under this rule,
$J(\pi)=\mathbb E[U_z(x,y,\asel)]$ is the selected-action score, while
$J_{\rm auth}(\pi)=\mathbb E[U_z(x,y,\hat a)]$ is the local-rule proposal score. On changed cases,
computing $J_{\rm auth}$ requires the score that the declined proposal would have received under
those same criteria. This declined-proposal score is not an estimate of the patient outcome that the
proposal would have caused. Their authority gap, $\Deltaauth=J_{\rm auth}-J$, isolates the score
change introduced by local selection. Positive values mean that proposal scoring is optimistic
relative to selection, and negative values mean that it is pessimistic.

\begin{proposition}[Authority-gap factorization]
\label{prop:authority-factorization}
Let $I=\mathbf 1[\hat a\neq\asel]$, $p=\Pr(I=1)$, and, for $p>0$,
$\delta=\mathbb E[U_z(x,y,\hat a)-U_z(x,y,\asel)\mid I=1]$. Under any fixed sampling and
weighting rule, $\Deltaauth=p\delta$. At $p=0$, $\Deltaauth=0$ and $\delta$ is undefined.
Consequently, for $p>0$, the proposal-change rate does not determine $\Deltaauth$ without an
assumption fixing $\delta$, and two local review processes can share $p$ while their
authority gaps differ in magnitude or sign.
\end{proposition}
Cases with $I=0$ contribute zero, so conditioning on $I=1$ yields the
factorization. Here $p$ is the proposal-change rate and $\delta$ the mean signed
score difference on changed units under the declared weighting rule. Because
$p$ leaves $\delta$ unconstrained, the same change rate can yield gaps of either
sign; for example, $p=0.30$ and $\delta=\pm 2$ give
$\Deltaauth=\pm 0.60$.

\paragraph{Record recoverability.}
We call a quantity \emph{directly recoverable} when the recorded fields suffice to compute it
without additional assumptions. A record containing both actions and the selected-action score
does not by itself recover $\Deltaauth$ when proposals change, because $U_z$ depends on locally
held $z$: when $z$ remains local, the evaluation record must additionally include the declined
proposal's locally computed score under the same criteria (Section~\ref{sec:application}).

\section{Experimental Setup}
\label{sec:setup}

We instantiate \sys{} in a controlled study: proposals come from frozen models scored
on real dermoscopy images, while local review and context are simulated under a study-defined
rule. We use the ISIC 2019 dermoscopy training set, comprising Memorial Sloan Kettering (MSK),
HAM10000, and BCN20000
sources~\cite{isic2019challenge,codella2017isic,tschandl2018ham10000,hernandezperez2024bcn20000}.
Diagnoses map to four study-defined risk levels, very low to high, and four ordinal routing
actions span observation to urgent specialist assessment (Supplementary Material gives the
mapping). Lesion-aware splitting with image fallback
produces $17{,}629/3{,}735/3{,}967$ train/validation/test images, and known-lesion or
exact-pHash groups do not cross splits (a broader, unenforced pHash diagnostic at Hamming
threshold 4 finds 122 cross-split near-duplicate components, detailed in the Supplementary
Material). Code and reproduction instructions are available at
\url{https://github.com/flint-xf-fan/AuthEval}.

Frozen MedGemma 1.5 4B assigns each risk continuation a mean token log-likelihood from the image
and proposer-visible age, sex, and anatomic site. A validation-trained balanced multinomial
mapping, the action readout, ranks the four actions from these
likelihoods~\cite{sellergren2026medgemma}. The readout was selected and locked before test
access; its test balanced accuracy is $0.420$. Earlier ISIC releases entered
MedGemma pretraining~\cite{sellergren2026medgemma}, so
``held out'' applies to readout fitting, not foundation-model pretraining.

Services may cap escalation (capacity), impose a safety floor (safety), or apply
both (mixed). Private factors can additionally remove or replace an admissible action
and rescale the scoring weights below. A class-balanced ridge metadata proxy, fitted on training
labels, supplies decision-time risk context from age, sex, and anatomic site. Four frozen
synthetic profiles set per-profile capacity and safety bounds, and the gate selects the first
admissible action from the ranked slate, never observing the test-case diagnosis, ideal action,
or utility score. The Supplementary Material gives the full profile assignment, private-factor
mechanics, selection rule, and prompt.

With $\ell(a),\ell(y)\in\{0,1,2,3\}$ the action and reference levels, the declared score is
$U_z(x,y,a)=-w_u(z)\,s_y\,[\ell(y)-\ell(a)]_+-w_o(z)\,[\ell(a)-\ell(y)]_+$, where
$[t]_+=\max(t,0)$ and $s_y$ is the component indexed by $\ell(y)$ in
$(s_0,s_1,s_2,s_3)=(0.5,1,2,4)$. The under- and over-escalation weights $w_u,w_o$ have base
values set per profile and are rescaled by the private factors above. Zero is best, and a
one-level over-escalation costs $0.71$--$3.45$ units versus $4.6$--$9.0$ for under-escalation
at high risk, so upward and downward substitutions are non-equivalent under
the declared static score. The Supplementary Material lists the per-profile weight pairs and
effective ranges.

Image-equal estimates weight images, and cluster-equal estimates weight $2{,}090$
known-lesion/exact-pHash clusters. Confidence intervals (CIs) use $5{,}000$ whole-cluster
bootstrap replicates and are conditional on the fitted readouts and frozen proxy, gate, utility,
and simulated context. For the mechanism comparison, cases, context, gate, scoring rule, and
metadata proxy remain fixed while the proposal mechanism varies: MedGemma is compared with a LLaVA-Med
option-likelihood control whose expected ordinal score is discretized by validation-fitted
thresholds (balanced accuracy $0.249$)~\cite{llavamed2023,microsoft2024llavamed15}, and a validation-thresholded operational proposal
derived from a supervised linear readout over LLaVA-Med's frozen CLIP vision tower ($0.546$
balanced accuracy)~\cite{radford2021clip}.

\section{Results}
\label{sec:results}

\paragraph{Proposal and selected-action scores diverge under local review.}
Under the mixed simulated local review process, the selected action differs from the
MedGemma-derived proposal on $47.0\%$ of test images. The proposal score $J_{\rm auth}=-2.94$
and selected-action score $J=-3.31$ give an authority gap $\Deltaauth=+0.374$ simulator units
(95\% CI $[+0.091,+0.663]$): under this declared score, proposal-only scoring is optimistic by
that amount. The proposal-change rate $p=0.470$ and mean score change $\delta=0.797$ give
$p\delta=0.374$ after rounding, consistent with Proposition~\ref{prop:authority-factorization}.
Scoring the declined proposal makes this comparison possible; the change rate remains
simulator-specific. Table~\ref{tab:workshop-results} summarizes this result and three checks.

\begin{table}[t]
\centering
\caption{Controlled authority gaps on the ISIC test split ($n=3{,}967$ images; 2,090 clusters; simulator units); positive values denote optimistic proposal scoring. The final row is the paired probe-minus-LLaVA-Med gap difference under fixed capacity. $p_{\rm img}$ is the image-level proposal-change rate and $\delta_{\rm img}$ the mean score change on changed images. Weighting and bootstrap CIs (brackets) are defined in Section~\ref{sec:setup}.}
\label{tab:workshop-results}
\footnotesize
\setlength{\tabcolsep}{2.0pt}
\renewcommand{\arraystretch}{0.72}
\begin{tabular*}{\linewidth}{@{\extracolsep{\fill}}llrrcc@{}}
\toprule
Quantity & Setting & $p_{\rm img}$ & $\delta_{\rm img}$ & \shortstack{Image\\estimate} & \shortstack{Cluster\\estimate} \\
\midrule
$\Delta_{\rm authority}$ & MedGemma / mixed & $0.470$ & $0.80$ & $+0.374$ & $-0.206$ \\
 & & & & {\footnotesize $[+0.091,+0.663]$} & {\footnotesize $[-0.373,-0.037]$} \\[0.5pt]
$\Delta_{\rm authority}$ & MedGemma / capacity & $0.341$ & $2.18$ & $+0.744$ & $+0.056$ \\
 & & & & {\footnotesize $[+0.491,+1.010]$} & {\footnotesize $[-0.107,+0.218]$} \\[0.5pt]
$\Delta_{\rm authority}$ & MedGemma / safety & $0.281$ & $-0.18$ & $-0.051$ & $-0.224$ \\
 & & & & {\footnotesize $[-0.262,+0.158]$} & {\footnotesize $[-0.358,-0.093]$} \\[0.5pt]
\midrule
Paired gap & Fixed capacity & -- & -- & $+0.810$ & $+0.433$ \\
 & & & & {\footnotesize $[+0.629,+0.999]$} & {\footnotesize $[+0.345,+0.524]$} \\[0.5pt]
\bottomrule
\end{tabular*}
\end{table}

\paragraph{Change frequency does not determine the authority gap.}
Capacity and safety yield proposal-change rates of $0.341$ and $0.281$, but mean score changes
of $2.18$ and $-0.18$ on changed cases. The corresponding image-equal authority gaps are
$+0.744$ ($[+0.491,+1.010]$) and $-0.051$ ($[-0.262,+0.158]$); capacity substitutions are
$97.6\%$ downward, whereas the safety arm is near zero. A rate-only adjustment with one shared
conditional value would give both gaps the same sign and scale them only by $p$. Instead, they
differ by $0.795$ units with opposite point-estimate signs, though only the capacity interval
excludes zero. Change frequency therefore determines neither size nor direction.

\paragraph{The gap varies across proposal mechanisms.}
Under a fixed capacity gate, the image-equal authority gap is $+0.014$ for a LLaVA-Med
option-likelihood head, $+0.744$ for MedGemma, and $+0.824$ for a linear probe over
LLaVA-Med's frozen CLIP vision tower. A paired cluster bootstrap places probe minus LLaVA-Med at $+0.810$ (95\% CI
$[+0.629,+0.999]$), and at $+0.433$ (95\% CI $[+0.345,+0.524]$) under cluster-equal weighting.
Because the mechanisms differ in architecture, readout, input signals, and slate construction,
this analysis supports neither component attribution nor a model-quality comparison; balanced
accuracy does not explain their gap differences. A capacity ceiling binds mechanisms in
proportion to how often they escalate through predominantly downward substitutions. Thus, a
gap for one complete proposal mechanism need not hold for another under the same review process.

\paragraph{What counts as one case changes the estimate.}
Giving each image equal weight produces the primary MedGemma authority gap $+0.374$, while
giving each lesion-aware cluster---images sharing a known lesion ID or exact perceptual-hash
match, with image fallback---equal weight produces $-0.206$
($[-0.373,-0.037]$). The two units differ because lesions are photographed unequally: a
high-risk cluster holds $2.98$ images on average against $1.54$ for a very-low-risk cluster.
Image weighting therefore emphasizes repeated high-risk lesions, precisely where a capacity
ceiling's downward substitution is most expensive under the asymmetric score. Cluster
weighting instead gives each lesion one vote.

Capacity changes from $+0.744$ to $+0.056$ ($[-0.107,+0.218]$) and safety from
$-0.051$ to $-0.224$ ($[-0.358,-0.093]$): under cluster weighting, the capacity estimate becomes
indistinguishable from zero while the safety estimate excludes zero. Together, these checks
show that proposal accuracy and change frequency cannot substitute for direct measurement of
the authority gap, and that the proposal mechanism and evaluation unit must be declared: \sys{}
enables that measurement from a full counterfactual record.

% \FloatBarrier
\section{Operationalizing \sys{} and Limitations}
\label{sec:application}

Operationally, the authority gap requires the proposed and selected actions, the selected-action score, and, when they differ, the declined proposal's score under the same local rule. Because that rule may depend on locally held context, the latter score must be computed at the site, either from a computable policy or retrospective adjudication that leaves the original selection unchanged. This keeps the underlying context loca~\cite{fan2025position}; if the score is unavailable, AuthEval reports the gap as not recoverable. The Supplementary Material gives the record levels and reporting rules.

Our results come from one imaging task in which the review process, the private
context, the workflow profiles, and the scoring weights are all study-defined, so they
establish that an authority gap can arise and change sign rather than estimating its size
in clinical practice. The intervals are conditional on those authored choices, and the
three proposal mechanisms differ along multiple dimensions. The Supplementary Material
details these boundaries and the evidence needed to extend the claims beyond them.

\section{Conclusion}

  Medical VLM benchmarks
  measure the quality of a model's proposed action, whereas clinical
  workflows also require a service to select an action under local capacity, policy,
  and patient constraints. Treating these as the same evaluation object can make
  workflow performance appear better or worse for reasons introduced after the
  model's proposal. We introduced \sys{} to keep these roles separate: it records
  the proposed and selected actions, evaluates both under the same local criteria,
  and expresses their difference as the authority gap. That gap is the product of
  the change rate and the average shift when it happens, so frequency alone
  cannot recover it. In our controlled ISIC study, its
  magnitude and direction varied across simulated review processes, proposal mechanisms, and
  evaluation units, and, under mixed constraints, it reversed sign when weighting
  shifted from image to lesion-aware cluster. \sys{} therefore lets studies making
  workflow-level claims test whether proposal scores remain representative after
  local review, measuring the resulting gap when local rescoring is available and
  stating clearly which workflow claims the available record can support.

% \begin{credits}
% % \subsubsection{\ackname} A bold run-in heading in small font size at the end of the paper is
% % used for general acknowledgments, for example: This study was funded
% % by X (grant number Y).

% \subsubsection{\discintname}
% The authors have no competing interests to declare.
% \end{credits}

\bibliographystyle{splncs04}
\bibliography{refs}




% LNCS Homepage, \url{http://www.springer.com/lncs}, last accessed 2023/10/25
% \end{thebibliography}

% ===== BEGIN SCIENTIFIC SUPPLEMENT =====
% Cut from this marker through the matching END marker to separate the supplement.
\clearpage
\appendix
\section{Scientific Supplement}

This supplement records the cohort construction, proposal mechanisms, simulated
review process, declared score, and supporting results used in the main paper.
All tables report frozen study artifacts; no additional model selection was
performed for the supplement. The frozen configurations, the retained per-case
traces, and a verifier that reproduces the reported values without ISIC images or model weights
are available in the reproduction package at \url{https://github.com/flint-xf-fan/AuthEval}.

\subsection{Scope and Limitations}\label{sec:supp-scope}

\sys{} is evaluated in a controlled study built from one public dermoscopy
collection. Proposals come from frozen models scored on real images, while local
review, private context, and the scoring weights follow a study-defined rule.
The evidence therefore characterizes an evaluation construct rather than
clinical prevalence, benefit, or harm. ISIC records no referral, capacity,
clinician-selection, completion, or outcome fields, so the workflow surrounding
each image is authored rather than observed.

The four workflow profiles are study assignments rather than institutions, and
the private factors are simulated independently of any recorded patient
attribute. The local review rules are stress tests chosen to span admissible
behavior, not clinical prescriptions, and the declared score is a static study
score rather than an outcome model. A gap measured under one such rule
constrains neither the gap under a different rule nor the clinical consequence
of the substitutions producing it.

The authority-gap endpoint and mechanism analyses reported here were conducted
post hoc after a prospectively locked policy-selection comparison yielded the
same policy under proposal-only and authority-aware objectives.

The three proposal mechanisms differ in architecture, readout, input signals,
and slate construction, so ordering them by authority gap does not isolate a single component:
three arms establish that the gap moves, not why it
moves. Image-equal weighting was designated primary after exploratory
development, and the cluster-equal estimand accompanies it throughout because
the sign of the primary estimate depends on that choice.

Stability after excluding the broader threshold-4 pHash components remains
untested because that heuristic can also connect visually similar but distinct
lesions.

Earlier ISIC releases entered MedGemma pretraining, so ``held out'' describes
the fitting of the action readout and the control thresholds rather than the
foundation model itself. Every estimate is conditional on the fitted readouts
and on the frozen proxy, gate, score, and simulated context, and the confidence
intervals propagate sampling variation across clusters alone.

\subsection{Cohorts, Splits, and Evaluation Units}

\paragraph{Risk strata and routing actions.}
The study uses the ISIC 2019 training metadata, ground truth, and image
release. Diagnoses map to four ordinal risk levels: very low (melanocytic nevus,
vascular lesion, or dermatofibroma), low (benign keratosis), moderate (actinic
keratosis or intraepithelial carcinoma, and basal-cell carcinoma), and high
(squamous-cell carcinoma or melanoma). Four ordinal routing actions
$a_0,\dots,a_3$ span observation, routine review, specialist assessment, and
urgent specialist assessment. The reference state $y$ is the risk level implied
by the recorded diagnosis and enters scoring only. Table~\ref{tab:supp-split}
reports the realized split composition.

\begin{table}[tb]
\centering
\caption{Split sizes and risk-stratum composition. Strata are derived from the
recorded diagnosis and are used only as the post-hoc reference state.}
\label{tab:supp-split}
\small
\begin{tabular}{@{}lrrrrr@{}}
\toprule
Split & $n$ & Very low & Low & Moderate & High \\
\midrule
Train & 17,629 & 9,392 & 1,839 & 2,890 & 3,508 \\
Validation & 3,735 & 1,923 & 393 & 648 & 771 \\
Test & 3,967 & 2,052 & 392 & 652 & 871 \\
\bottomrule
\end{tabular}
\end{table}

\paragraph{Split construction and duplicate control.}
Images sharing a known lesion identifier or an exact 64-bit discrete cosine transform
(DCT) perceptual hash (pHash) are assigned to the same split, so neither relation crosses the
train, validation, and test boundaries. A broader pHash diagnostic at Hamming
threshold 4 is computed for reference but does not redefine the split, because
that threshold also connects visually similar yet distinct lesions.
Table~\ref{tab:supp-duplicates} summarizes both groupings. The
122 components that cross splits under the broader threshold are the reason the
diagnostic is reported rather than enforced.

\begin{table}[tb]
\centering
\caption{Perceptual-duplicate groupings over all 25,331 images. Exact groups
constrain the split; threshold-4 components are diagnostic only.}
\label{tab:supp-duplicates}
\small
\begin{tabular}{@{}lrr@{}}
\toprule
Grouping & Components & Images \\
\midrule
Exact pHash & 95 & 191 \\
Exact pHash crossing splits & 0 & 0 \\
Threshold-4 pHash & 594 & 1,777 \\
Threshold-4 pHash crossing splits & 122 & --- \\
\bottomrule
\end{tabular}
\end{table}

\paragraph{Evaluation units.}
The primary estimand weights the 3,967 test images equally. The sensitivity
estimand weights 2,090 clusters equally, of which 1,778 derive from known lesion
identifiers and 312 from images carrying none, after exact-hash unions. Each
cluster receives one simulated context, so its members share a workflow profile,
private factors, and mean metadata proxy score. Confidence intervals resample whole clusters, using 5,000
replicates and the 0.025 and 0.975 quantiles.

The two units are not interchangeable, because lesions are photographed unequally
across the risk strata. Table~\ref{tab:supp-clustersize} gives the imbalance:
high-risk clusters carry roughly twice as many images as very-low-risk clusters, so
image-equal weighting places proportionally more weight on high-risk lesions than
cluster-equal weighting does. This is the mechanism behind the sign reversal
reported in the main paper.

\begin{table}[tb]
\centering
\caption{Test-split cluster sizes by risk stratum. Image-equal weighting is
proportional to the image column, cluster-equal weighting to the cluster column.}
\label{tab:supp-clustersize}
\small
\begin{tabular}{@{}lrrr@{}}
\toprule
Risk stratum & Clusters & Images & Images per cluster \\
\midrule
Very low & 1,333 & 2,052 & 1.54 \\
Low & 221 & 392 & 1.77 \\
Moderate & 244 & 652 & 2.67 \\
High & 292 & 871 & 2.98 \\
\bottomrule
\end{tabular}
\end{table}

\subsection{Proposal Mechanisms}

\paragraph{MedGemma proposal.}
We run the primary proposer, MedGemma 1.5 4B IT, in bfloat16 at the frozen
identifier and revision:
\begin{center}
\texttt{google/medgemma-1.5-4b-it}\\
\texttt{91850547d9f0b2fdd21aa7c5f4f3d1a8a52c243b}
\end{center}
For each image and the proposer-visible age, sex, and
anatomic site, the frozen model assigns a mean token log-likelihood to four
semantic continuations under the prompt:

\begin{quote}\small
This is a research image-assessment task, not patient-specific clinical advice.
Inspect the dermoscopy image and the listed public metadata only. Rank the
lesion into one of four ordinal visual-risk categories. Do not infer unavailable
history, symptoms, evolution, patient preference, local policy, or service
capacity. Categories: very low visual risk; low visual risk; moderate visual
risk; high visual risk. Public metadata:\\
\texttt{age=\{age\}, sex=\{sex\},}\\
\texttt{anatomic\_site=\{anatomic\_site\}}. Selected visual-risk category:
\end{quote}

The four continuations describe, respectively, a benign appearance without
evident suspicious dermoscopic features; a mostly benign appearance with limited
atypical features; suspicious features warranting specialist assessment; and
strongly suspicious features warranting urgent specialist assessment.

\paragraph{Action readout.}
A balanced multinomial logistic readout maps the four log-clipped continuation
probabilities (clip $10^{-8}$) to the four routing actions. Four-fold grouped
validation searches $C\in\{0.01,0.1,1,10\}$, minimizing ordinal mean absolute
error and then maximizing balanced accuracy and macro-F1, with smaller $C$ as
the final tie-break; the selected value is $C=0.01$. Test labels enter neither
fitting nor selection. On the held-out validation set, the readout uses all four
actions and improves ordinal MAE from $1.104$ for the best constant to $0.860$.
Shuffling images against their metadata within a 256-image validation subset
raises MAE by $0.375$ and lowers balanced accuracy by $0.132$, indicating that
the readout responds to image content rather than metadata alone.

\paragraph{Proposal controls.}
Two frozen controls supply alternative proposals through the same interface:
\begin{center}
\texttt{microsoft/llava-med-v1.5-mistral-7b}\\
\texttt{openai/clip-vit-large-patch14-336}
\end{center}
The language control reads option likelihoods from the first and converts its
expected ordinal score to an action; the vision control uses a balanced linear
probe trained on frozen CLIP ViT-L/14 features from the second. Both discretize with
validation-fitted thresholds:
\[
\begin{aligned}
\text{language control:}\quad & (0.602433,\ 0.645141,\ 0.660500),\\
\text{vision control:}\quad & (0.459906,\ 0.551713,\ 0.662236).
\end{aligned}
\]
Both controls are consumed by the simulator only through these frozen operational
heads, each a function of the image and declared public metadata alone.

\subsection{Simulated Review Process}

\paragraph{Workflow profiles.}
Each cluster is assigned one of four workflow profiles by a stable hash of its
age band and anatomic site, falling back to the image identifier only when both
fields are missing; a cluster takes the modal profile of its members. A profile
fixes an upper capacity bound on escalation, a floor policy, and the base
under- and over-escalation weights of Table~\ref{tab:supp-profiles}. The
profiles are labeled by the service character they are meant to span, from a
high-capacity dermatology service to a low-capacity, access-constrained one.

\begin{table}[h!]
\centering
\caption{Workflow-profile parameters. Capacity is the highest admissible action
level; $w_u$ and $w_o$ are the base under- and over-escalation weights before
the private multipliers of Table~\ref{tab:supp-factors}.}
\label{tab:supp-profiles}
\small
\begin{tabular}{@{}llccrr@{}}
\toprule
Profile & Service character & Capacity & Floor policy & $w_u$ & $w_o$ \\
\midrule
P1 & High-capacity dermatology & 3 & Standard & 1.20 & 0.75 \\
P2 & General mixed source & 2 & Standard & 1.20 & 1.20 \\
P3 & Specialist, safety-oriented & 3 & Strict & 1.80 & 0.90 \\
P4 & Low-capacity, access-constrained & 1 & Lenient & 1.30 & 2.00 \\
\bottomrule
\end{tabular}
\end{table}

\paragraph{Private factors.}
Within each cluster, one level of each factor in
Table~\ref{tab:supp-factors} is drawn and held fixed. Sensitivity preference
rescales both weights; follow-up burden and invasiveness aversion rescale the
over-escalation weight alone. Invasiveness aversion additionally removes $a_3$
from the admissible set, and high follow-up burden replaces an admissible $a_1$
by $a_0$ in lower proxy bins or by $a_2$ in higher bins when that alternative is
itself admissible.

\begin{table}[tb]
\centering
\caption{Private factor levels and their multipliers on the under- and
over-escalation weights.}
\label{tab:supp-factors}
\small
\setlength{\tabcolsep}{5.0pt}
\begin{tabular}{@{}llrr@{}}
\toprule
Factor & Level & $w_u$ mult. & $w_o$ mult. \\
\midrule
Sensitivity preference & Sensitivity-seeking & 1.25 & 0.95 \\
 & Neutral & 1.00 & 1.00 \\
 & Specificity-seeking & 0.95 & 1.25 \\
\midrule
Follow-up burden & Low & 1.00 & 1.00 \\
 & High & 1.00 & 1.15 \\
\midrule
Invasiveness aversion & Neutral & 1.00 & 1.00 \\
 & Averse & 1.00 & 1.20 \\
\bottomrule
\end{tabular}
\end{table}

\paragraph{Gate proxy and selection rule.}
The decision-time risk context is a class-balanced ridge model fitted on
training labels from age, sex, and anatomic site, with regularization and
temperature selected on validation. Its score is binned at $(0.25, 0.50, 0.75)$.
The profile supplies the capacity bound, the proxy bin together with the profile
floor policy supplies the safety floor, and the mixed regime intersects both.
When an override empties the admissible band, the declared default is $a_0$ in
lower proxy bins and $a_2$ in higher bins. The gate selects the first admissible
action in the ranked slate and never observes the diagnosis, the reference
action, or the score, so selection cannot depend on the evaluation target.

\subsection{Declared Score}

\paragraph{Score definition.}
For action and reference levels $\ell(a),\ell(y)\in\{0,1,2,3\}$, the declared
score is
\begin{equation}
U_z(x,y,a)=-w_u(z)\,s_y\,[\ell(y)-\ell(a)]_+-w_o(z)\,[\ell(a)-\ell(y)]_+,
\label{eq:supp-score}
\end{equation}
with $[t]_+=\max(t,0)$ and $s_y$ the component indexed by $\ell(y)$ in
$(s_0,s_1,s_2,s_3)=(0.5,1,2,4)$. Zero is the best attainable value.

\paragraph{Weight ranges and asymmetry.}
Applying the multipliers of Table~\ref{tab:supp-factors} to the base weights of
Table~\ref{tab:supp-profiles} gives effective ranges $w_u\in[1.14,2.25]$ and
$w_o\in[0.71,3.45]$. A one-level over-escalation therefore costs between
$0.71$ and $3.45$ units, whereas a one-level under-escalation at high risk
costs between $4.6$ and $9.0$, since $s_3=4$ multiplies the under-escalation
weight. Upward and downward substitutions are consequently not equivalent under
Eq.~\eqref{eq:supp-score}, contributing to the different gap magnitudes and signs
produced by the two review processes.

\paragraph{Relation to a benchmark metric.}
A benchmark may report a score $J_{\rm bench}$ under a different metric. When
$J_{\rm bench}$ is commensurate with $U_z$ and uses the same cases, evaluation
distribution, and weighting rule, its difference from the selected-action score
decomposes as
\begin{equation}
J_{\rm bench}-J=(J_{\rm bench}-J_{\rm auth})+\Deltaauth ,
\label{eq:supp-decomposition}
\end{equation}
where the first term is the gap between the benchmark metric and the declared
local score evaluated on the same proposal, and the second is the local-selection
effect isolated in the main paper. A comparison that does not hold the first
term fixed cannot attribute a change in $J_{\rm bench}-J$ to local selection
alone.

\subsection{Record Levels}

Table~\ref{tab:supp-records} states which \sys{} quantities each cumulative record level
supports. The levels are cumulative: each adds the fields shown to the preceding level and
thereby supports the quantities listed without additional assumptions.

Every record also versions the case and cluster identifiers, proposal interface, local
selection and scoring rules, reference state, and evaluation unit, and may log actor,
reason, timestamp, and completion as further workflow metadata. If $p=0$, \sys{} reports
$\Deltaauth=0$ and leaves $\delta$ undefined; if $p>0$ but the declined proposal's local
score is unavailable, it reports $\Deltaauth$ and $\delta$ as not recoverable. Exporting
locally computed scores rather than raw context can reduce data transfer, but provides no
formal privacy guarantee.

\begin{table}[h!]
\centering
\caption{Cumulative record levels and the \sys{} quantities each one recovers. Fields are
additional to the level above.}
\label{tab:supp-records}
\small
\begin{tabular}{@{}lll@{}}
\toprule
Record level & Additional local-review fields & Quantities recovered \\
\midrule
Proposal-only & $\hat a$ and proposal metric & none \\
Role-separated & $\asel$, $U_z(x,y,\asel)$ & $p$, $J$ \\
Full counterfactual & $U_z(x,y,\hat a)$ on changed cases
& $J_{\rm auth}$, $\Deltaauth$, $\delta$ if $p>0$ \\
\bottomrule
\end{tabular}
\end{table}

\subsection{Supporting Results}

\paragraph{Review-process decomposition.}
Table~\ref{tab:supp-regimes} reports the two constituent review processes
alongside the mixed process of the main paper. The proposal score $J_{\rm auth}$
is identical across the three rows because the proposal does not depend on the
gate; only the selected-action score $J$ moves. The capacity ceiling constrains
actions through the private profile and factors, and the safety floor constrains
them through the public metadata proxy that the proposer also sees. Under image
weighting the discrepancy is carried by the capacity channel, whose
substitutions are $97.6\%$ downward, while the safety channel is
indistinguishable from zero; under cluster weighting the ordering reverses, and
the safety channel is the one whose interval excludes zero.

\begin{table}[tb]
\centering
\caption{Authority gap by simulated review process for the fixed
MedGemma-derived proposal. Scores are in simulator units; brackets give 95\%
whole-cluster bootstrap intervals.}
\label{tab:supp-regimes}
\small
\setlength{\tabcolsep}{3.0pt}
\resizebox{\linewidth}{!}{%
\begin{tabular}{@{}lrrrrll@{}}
\toprule
Review process & $J_{\rm auth}$ & $J$ & $p$ & $\delta$
& $\Deltaauth$, image & $\Deltaauth$, cluster \\
\midrule
Capacity ceiling & $-2.938$ & $-3.683$ & $0.341$ & $2.182$
& $+0.744$ $[+0.491,+1.010]$ & $+0.056$ $[-0.107,+0.218]$ \\
Safety floor & $-2.938$ & $-2.888$ & $0.281$ & $-0.180$
& $-0.051$ $[-0.262,+0.158]$ & $-0.224$ $[-0.358,-0.093]$ \\
Mixed & $-2.938$ & $-3.313$ & $0.470$ & $0.797$
& $+0.374$ $[+0.091,+0.663]$ & $-0.206$ $[-0.373,-0.037]$ \\
\bottomrule
\end{tabular}}
\end{table}

\paragraph{Proposal dependence.}
Table~\ref{tab:supp-proposals} holds the capacity gate, cases, private context,
clusters, and score fixed while varying the proposal mechanism. Both the change rate
and the conditional score change vary with the mechanism; under cluster weighting,
the language control and vision probe have authority gaps of opposite sign. The paired
contrast evaluates both proposals on the same clusters within each replicate, and its
confidence interval excludes zero under both evaluation units.

\begin{table}[h!]
\centering
\caption{Authority gap by proposal mechanism under one fixed capacity review
process. The final row is the paired probe-minus-LLaVA-Med contrast, computed on
identical clusters within each bootstrap replicate.}
\label{tab:supp-proposals}
\small
\setlength{\tabcolsep}{3.0pt}
\resizebox{\linewidth}{!}{%
\begin{tabular}{@{}lrrrll@{}}
\toprule
Proposal & Test balanced acc. & $p$ & $\delta$
& $\Deltaauth$, image & $\Deltaauth$, cluster \\
\midrule
LLaVA-Med head & $0.249$ & $0.163$ & $0.084$
& $+0.014$ $[-0.056,+0.086]$ & $-0.096$ $[-0.156,-0.037]$ \\
MedGemma readout & $0.420$ & $0.341$ & $2.182$
& $+0.744$ $[+0.491,+1.010]$ & $+0.056$ $[-0.107,+0.218]$ \\
CLIP probe & $0.546$ & $0.276$ & $2.988$
& $+0.824$ $[+0.621,+1.033]$ & $+0.337$ $[+0.256,+0.422]$ \\
\midrule
Probe $-$ LLaVA-Med & --- & --- & ---
& $+0.810$ $[+0.629,+0.999]$ & $+0.433$ $[+0.345,+0.524]$ \\
\bottomrule
\end{tabular}}
\end{table}

% ===== END SCIENTIFIC SUPPLEMENT =====

\end{document}